\documentclass[twocolumn]{article}

\usepackage{cite}
\usepackage{array}
\usepackage{algorithm}
\usepackage{algorithmic}
\usepackage{amsmath,amssymb,amsfonts}
\usepackage{graphicx}
\usepackage{authblk}
\usepackage{textcomp}
\usepackage{mathrsfs}
\usepackage{xcolor}
\usepackage{hyperref}
\usepackage{multirow}
\usepackage{url}

\def\BibTeX{{\rm B\kern-.05em{\sc i\kern-.025em b}\kern-.08em
    T\kern-.1667em\lower.7ex\hbox{E}\kern-.125emX}}

\date{}

\begin{document}
\title{Compression of Recurrent Neural Networks using Matrix Factorization}

\author[1,2,3,*]{Lucas Maison}
\author[1]{Hélion du Mas des Bourboux}
\author[1]{Thomas Courtat}

\affil[1]{Thales SIX, ThereSIS ML and Datascience Lab, Palaiseau, France}
\affil[2]{Laboratoire Informatique d'Avignon, Avignon, France}
\affil[3]{Thales SIX, HTE Multimedia Lab, Gennevilliers, France}
\affil[*]{Corresponding author: Lucas Maison, lucas.maison@thalesgroup.com}

\maketitle
\begin{abstract}
Compressing neural networks is a key step when deploying models for real-time or embedded applications. Factorizing model's matrices using low-rank approximations is a promising method for achieving compression. While it is possible to set the rank before training, this approach is neither flexible nor optimal. In this work, we propose a post-training rank-selection method called Rank-Tuning that select a different rank for each matrix. Used in combination with training adaptations, our method achieve high compression rates with no or little performance degradation. Our numerical experiments on signal processing tasks show that we can compress recurrent neural networks up to $14\times$ with at most $1.4\%$ of relative performance reduction.\footnote{Code and models are publicly available at \url{https://github.com/Deathekirl/low-rank-approximation}, together with instructions for downloading the datasets.}
\end{abstract}

\providecommand{\keywords}[1]
{
  \small	
  \textbf{\textit{Keywords---}} #1
}


\message{The column width is: \the\columnwidth}

\section{Introduction and Related Work}

We observe a clear tendency of neural network size growth~\cite{DNN_size}. New architectures using hundreds of millions or even billions of parameters are elaborated, leading to state-of-the-art performance but also to increasing difficulties: slowness of inference, increased demand for storage, exploding electricity consumption and pollution. These issues inhibit the deployment of high performance neural network models in real-time and embedded applications.

Several general strategies for reducing networks' complexity have been explored with success, including distillation~\cite{hinton2015distilling}, pruning~\cite{lecun1990optimal}, quantization~\cite{vanhoucke2011improving}, low-rank approximations~\cite{denil2013predicting}. The interested reader can refer to~\cite{mishra2020survey} for a survey of network compression methods. In this work we focus on low-rank approximations which are a promising technique not yet included in deep learning libraries at the time of writing.

There exist several methods for approximating matrices, for instance the Low-Rank (LR) decomposition~\cite{denil2013predicting} where Singular Value Decomposition (SVD) is used to factorize matrices, reducing the number of operations to carry out for the matrix product. Another example is the Tensor-Train (TT) decomposition~\cite{grachev2019compression} where a matrix is represented as a product of several tensors.

The LR decomposition can be exploited in a number of ways. The most common approach is to learn the weights of pre-factorized matrices by fixing their rank before training~\cite{grachev2019compression,lu2016learning,davis2013low,sainath2013low,denil2013predicting}. However, this strategy requires a costly\footnote{It is necessary to re-train the model from scratch for each ranks combination.} research of optimal ranks, or a-priori knowledge of them. Moreover, it's not very adaptable since ranks cannot be changed after training.

Another approach consists in learning whole matrices and factorize them after training. There exist several training adaptations that can greatly enhance the quality of the decomposition. In~\cite{xu2020trp,yaguchi2019decomposable,alvarez2017compression}, authors apply a regularization based on matrix norms (see section~\ref{subsubsec:regularization}). Trained Rank Pruning (TRP,~\cite{xu2020trp}) combines regularization and training-time factorization (see section~\ref{subsubsec:hlra}) and uses \textit{energy} as a criterion for choosing ranks. The Learning-Compression (LC,~\cite{idelbayev2020low}) algorithm enable learning weights and ranks alternatively.

These methods have various application areas. One can factorize deep neural networks for vision tasks \cite{davis2013low,denil2013predicting} or ASR tasks \cite{sainath2013low}. In~\cite{idelbayev2020low,xu2020trp,yaguchi2019decomposable,alvarez2017compression,denil2013predicting}, authors factorize convolutional networks (LeNet, VGG, ResNet) evaluated on image recognition tasks (MNIST, CIFAR, ImageNet, etc.). In~\cite{grachev2019compression,lu2016learning}, authors factorize recurrent networks (RNN, LSTM) evaluated on ASR (Automatic Speech Recognition) or linguistic modeling tasks.

Regularization and training-time factorization have been used in \cite{xu2020trp} to compress convolutional networks. Recurrent networks have also been compressed using ranks fixed prior to training \cite{lu2016learning}.

The present paper aims at compressing recurrent neural networks through training-time factorization and regularization. To the best of our knowledge, it is the first work implementing both these strategies while also selecting the best ranks during post-training.

The rest of this article is organized as follows. In section \ref{sec:math_bg} we first introduce the mathematical background of our work, then present the two training adaptation methods we used, and finally describe post-training rank selection strategies. In section \ref{sec:experiments} we present the different evaluation tasks and our model's architecture. We present our results and discuss our findings in section \ref{sec:results}. Section \ref{sec:conclusion} concludes this study.
\section{Low-Rank approximations for neural networks }
\label{sec:math_bg}

This section introduces the key concept of SVD. Then we present two techniques used at training time to obtain matrices of low-rank and achieve better compression. Finally, several post-training rank selection strategies are presented.

\subsection{Using SVD to factorize a model}
\label{subsec:svd_basics}

A neural network $\mathcal{M}$ with parameters $\theta_{\mathcal{M}}$ is composed of an ensemble of $M$ weight matrices and bias vectors. For instance, a linear layer uses a matrix $W$ and a vector $b$ to perform the operation $y=Wx+b$. Same reasoning can be applied to more complex units like a GRU (Gated Recurrent Unit~\cite{gru}) or an attention layer used in the Transformer~\cite{Vaswani2017} architecture, which are decomposable in simple matrix-vector operations. The size of the vectors being negligible when compared to the size of the matrices, we focus on compressing the latter.

Let's write a weight matrix $W \in M(n,m)$ and a data matrix $A \in M(m, k)$. It costs $nm$ memory units to store $W$, and $nmk$ operations to compute $W \times A$. It is possible to decompose $W$ as a product of three matrices
\begin{equation}\label{eq:svd_definition}W = U\Sigma V^T = U\textrm{diag}(\sigma_1,...,\sigma_m)V^T,\end{equation}
with $U \in M(n,n)$, $\Sigma \in M(n,m)$ and $V \in M(m,m)$. Note that $\Sigma$ contains the \textit{singular values} of $W$, sorted in descending order. According to the Eckart-Young theorem~\cite{eckart1936approximation}, this is an accurate decomposition that always exists.

However, it is also possible to approximate this factorization if we truncate the $\Sigma$ matrix, keeping only its largest values. The factorization is said to be of \textit{rank} $r$ if we keep the $r$ first singular values and drop the rest. In this case we have
\begin{equation}W \simeq \hat{W} = U_r \Sigma_r V_r^T = (U_r\Sigma_r)V_r^T = \hat{U}_r V_r^T,\end{equation}
with $r \leq \min(n,m)$ and $U_r$, $\Sigma_r$, $V_r^T$ = $U[:, \mathpunct{:} r]$, $\Sigma[\mathpunct{:} r]$, $V^T[\mathpunct{:} r, :]$.
As we increase the rank $r$, the quality of the approximation improves~\cite{eckart1936approximation}, as illustrated by the formula
\begin{equation}\label{eq:norm_error}\left\|W - \hat{U}_r V_r^T \right\|_F = \sqrt{\sum_{i=r+1}^m \sigma_i^2}.\end{equation}

The storage cost of $\hat{W}$ is $r(n+m)$, and the matrix multiplication $W \times A$ becomes
\begin{equation}\label{eq:matrix_multi}WA \simeq \hat{W}A = \hat{U}_r V_r^TA = \hat{U}_r(V_r^TA),\end{equation}
which costs $r(n+m)k$ operations. In the end, we obtain both a memory and complexity gain if the following condition is met:
\begin{equation}r < \frac{nm}{n+m}.\label{eq:threshold_rank}\end{equation}
It is therefore possible to both compress the model and accelerate its inference by decomposing the weight matrices using low-ranks.\footnote{In theory, there is a speed gain as soon as equation \ref{eq:threshold_rank} is verified. In practice, the gain depends on the compute library implementation.} The lower the ranks, the better the compression. However, this could lead to a precision diminution if the matrices are too much approximated. We will explore this trade-off in details in section \ref{subsec:trade-off}.

\subsection{Training adaptations for neural networks compression}

SVD can be leveraged to compress any model, including those already trained. However, the compression rate may be poor if no adaptation has been used during training, as we will see in section~\ref{sec:results}. Here, we present two complementary ways of adapting the training which will increase the compression potential of the model without hurting much its performance.

\subsubsection{Nuclear Regularization}
\label{subsubsec:regularization}

The rank of a matrix $W$ is
\begin{equation}\textrm{rank}(W) = \sum_{i=1}^m \mathbf{1}_{\sigma_i > 0}.\label{eq:rank_definition}\end{equation}
\noindent Minimizing the rank of $W$ is therefore equivalent to maximizing the number of null singular values. Note that if some singular values are sufficiently close to zero, setting them to zero will reduce the rank of the matrix without changing much its action on vectors. It follows that diminishing the singular values during training will reduce the rank of the matrices, making it easier to compress the model.

One method to do so is to add regularization to the loss function, effectively constraining the rank of the matrices during training. In this work we use the \textit{nuclear norm}\footnote{Also known as \textit{trace norm} or \textit{Schatten 1-norm}.} as a regularizer:
\begin{equation}\left\| W\right\|_{*} = \sum_{i=1}^m \sigma_i.\end{equation}

\noindent The loss at epoch $t$ then becomes
\begin{equation}\mathscr{L} = \sum_{(X,y)}\mathscr{L}_{task}(\mathcal{M}(X,\theta_{\mathcal{M}}),y) + \lambda(t)\sum_{W \in \theta_{\mathcal{M}}} \left\| W\right\|_{*},\end{equation}
where the sum is evaluated on all the network' weight matrices.\footnote{However, we decided to only take into account the GRU' matrices, since they represent more than 99\% of the network' weights.} $\mathscr{L}_{task}$ is the task's objective loss (e.g. cross-entropy for classification tasks and Mean Squared Error (MSE) for regression tasks), and $\lambda(t)$ is an ad-hoc time-depending function defined as
\begin{equation}\label{eq:lambda_definition}\lambda(t) = \bar{\lambda}\times
\begin{cases}
 0 & \text{ if } t < T_1 \\
 \frac{t - T_1}{T_2 - T_1} & \text{ if } T_1 \leq t < T_2 \\
 1 & \text{ otherwise }
\end{cases},\end{equation}
where $\bar{\lambda}$ is the final weight of the regularization and $T_1$ and $T_2$ are two pivot epoch numbers. This function allows for a progressive addition of the regularization during training. Indeed, we find that a model cannot learn if regularization is added from the first epoch. We believe that it needs some time to adjust its weights (which are randomly initialized), after which we can add regularization. 

\subsubsection{Hard Low-Rank Approximation}
\label{subsubsec:hlra}

Another method for reducing the ranks during training consists in factorizing the matrices every $N$ epochs with target rank $R$ (see algorithm \ref{alg:hlra} for details). We call this method \textit{Hard Low-Rank Approximation} (HLRA) because it imposes a brutal constraint on the network, effectively reducing the ranks at the cost of performance (i.e. the optimized metric). During training the model eventually regain the lost performance between the factorization steps. In our experiments, we obtain a model with low-rank matrices with no or little performance loss.

\begin{algorithm}[H]
    \caption{HLRA method}
    \label{alg:hlra}
	\begin{algorithmic}
        \REQUIRE Target rank $R$, period $N$
        \IF{$t \equiv 0 \mod{N}$}
            \FOR{$W_i \in \theta_{\mathcal{M}}$}
                \STATE $U,\Sigma,V^T = \textrm{SVD}(W_i)$
                \STATE $U_R, \Sigma_R, V_R^T = U[:, \mathpunct{:} R], \Sigma[\mathpunct{:} R], V^T[\mathpunct{:} R, :]$
                \STATE $\hat{W_i} = U_R \times \Sigma_R \times V_R^T$
            \ENDFOR
            \STATE $\theta_{\mathcal{M}} \leftarrow \{ \hat{W_1}, ..., \hat{W_M}\}$
        \ENDIF
	\end{algorithmic}
\end{algorithm}

\subsection{Post-training rank selection strategies}
\label{subsec:post_training}

After training, the next step is to factorize each matrix to its optimal rank\footnote{In some cases, it may be beneficial to not factorize the matrix at all: the matrix is then said to be \textit{full-rank}.} with respect to the model's performance (evaluated on the validation split). The optimal rank of each weight matrix is unknown. We could brute-force search the ranks, but this is intractable in practice. It is therefore necessary to use heuristics.\\

\subsubsection{Simple strategies}
\label{subsubsec:simple_strategies}

A naive approach for choosing ranks consists in setting all matrices to the same rank $R_0$ and then evaluating the model's performance.\footnote{Since matrices differ in size, some ranks may be too large for some of them. It is for instance impossible to decompose a $(300,200)$-matrix to the rank 250, since the rank has to be greater than both dimensions. In this case, the matrix is not decomposed.} We can choose the smallest $R_0$ such that the performance stays greater than some threshold.

This strategy however is likely not optimal, since some matrices in the network are harder to factorize than others. A small number of matrices will be responsible for a large part of the performance degradation when $R_0$ is low. One can easily spot these on figure~\ref{fig:singular_values}: the matrices of the deepest layers have slowly decreasing singular values. When ignoring hard-to-factorize matrices, we can obtain better compression results. This second approach is said to be \textit{adaptative} since, for a given rank, matrices are factorized only if the associated decomposition error $\left\|W - \hat{U}_r V_r^T \right\|_F$ is less than a given threshold, effectively filtering hard-to-factorize matrices.\\

\subsubsection{Rank-Tuning}
\label{subsubsec:rank_tuning}

We introduce Rank-Tuning, which is a more sophisticated method for selecting the ranks. It is described in algorithm~\ref{alg:rank_tuning}. We factorize each matrix $W_i$ it at rank $r_i$ while leaving all the others unfactorized. We then run the model with the compressed matrix and evaluate its performance $p$ on the validation dataset. If $p$ is at least at $\Delta_p$ of the topline performance $p^{*}$, we stop and move on to the next matrix. Otherwise, we keep increasing $r_i$ (therefore diminishing the approximation error) until the condition is met. Finally, when all the ranks $\hat{r_1}, ..., \hat{r_M}$ have been found, we factorize each matrix $W_i$ at rank $\hat{r_i}$ to obtain the compressed network.

\begin{algorithm}[h!]
	\caption{Rank-Tuning}
	\label{alg:rank_tuning}
	\begin{algorithmic}
        \REQUIRE Topline $p^{*}$, degradation tolerance $\Delta_p$
		\FOR{$W_i \in \theta_\mathcal{M}$}
            \STATE $n, m = \textrm{shape}(W_i)$
            \STATE $U,\Sigma,V^T = \textrm{SVD}(W_i)$
			\FOR{$1 \leq r < \frac{nm}{n+m}$}
                \STATE $U_r, \Sigma_r, V_r^T = U[:, \mathpunct{:} r], \Sigma[\mathpunct{:} r], V^T[\mathpunct{:} r, :]$
				\STATE $\hat{W_i} = U_r \times \Sigma_r \times V_r^T$
				\STATE $\hat{\theta}_\mathcal{M} \leftarrow \{ W_1, ..., \hat{W_i}, ..., W_M\}$
                \STATE $p = \textrm{performance}(\mathcal{M}, \hat{\theta}_\mathcal{M})$
                \IF{$p > p^* - \Delta_p$}
                	\STATE $\hat{r_i} \leftarrow r$
                	\STATE break
                \ENDIF
			\ENDFOR
		\ENDFOR
		\RETURN{$\hat{r_1}, ..., \hat{r_M}$}
	\end{algorithmic}
\end{algorithm}

\subsubsection{Complexity considerations}

The exact computational cost of the procedures defined in sections~\ref{subsubsec:simple_strategies} and~\ref{subsubsec:rank_tuning} is quite difficult to evaluate, for several reasons:
\begin{itemize}
    \item they act on matrices of various shapes;
    \item they rely on highly optimized operations (SVD, matrices operations) which are implemented by linear algebra libraries and whose complexities are not naive.
\end{itemize}

\noindent However, we remark that network's inference is the most costly operation performed. To get an idea of the computational cost of selecting the ranks, we can simply count the number of times we run inference on the validation set.

When using the simple strategies, we run one inference for each possible value of $R_0$. The number of inferences is of the order of $\mathcal{O}(\underset{n,m=\textrm{shape}(W_i)}{\max}(\min(n,m)))$. In our setting, this simplifies to $\mathcal{O}(h)$, where $h$ is the hidden size of the GRUs. See section~\ref{subsec:architecture} for more details on our models' architecture.

When using Rank-Tuning, we run one inference for each possible rank for each matrix. The number of inferences is of the order of $\mathcal{O}(\underset{n,m=\textrm{shape}(W_i)}{\sum}\frac{nm}{n+m})$ which simplifies to $\mathcal{O}(M\times h)$. The cost of running Rank-Tuning is noticeably higher than those of naive strategies, but it is likely to lead to better compression rates. Moreover, it is still far more efficient than a brute-force search, which would require $\mathcal{O}(h^M)$ inferences to run.
\section{Experimental Setup}
\label{sec:experiments}

In this section we first introduce the three tasks on public datasets we used for evaluating our method. We also describe the architecture of our model.

\subsection{Introducing the tasks}

\textbf{SequentialMNIST} is a toy dataset based on the MNIST database (Modified National Institute of Standards and Technology database~\cite{mnist}). It consists of a sequential representation of MNIST images: each $28 \times 28$ image is flattened, leading to a sequence of length 784, which is then divided in two sequences of equal length, representing the top and the bottom of the image.

Given the first sequence (top), the task is to reconstruct the second (bottom). This is a sequence-to-sequence problem operating on vectors of size 392. We use the original train/test splits, and 20\% of the training set for validation during training. The MSE between the output sequence and the target sequence is used as the learning loss and as the performance metric at test time.\\
 
\noindent\textbf{DOCC10} (Dyni Odontocete Click Classification, 10 species~\cite{DOCC10}) is a dataset of marine mammals echolocation clicks. Each sample is a sequence of length 8192 centered on a \textit{click} of interest. The task is to classify each sample in one of the 10 classes of marine mammals. The dataset is balanced and composed of $134\,080$ samples, with the train/test splits available on the online challenge website\footnote{\url{https://challengedata.ens.fr/participants/challenges/32/}}. We use 20\% of the training set for validation during training. Cross-Entropy is used as the learning loss, while the test performance metric is classification accuracy.\\

\noindent\textbf{AugMod}\cite{AugMod}, which stands for \textit{Augmented Modulation}, is a synthetic dataset of radio signals. It is publicly available.\footnote{\url{https://www.kaggle.com/datasets/hdumasde/pythagoremodreco}} The authors used seven different linear modulations and five bins of SNR~(Signal to Noise Ratio). They generated 5000 examples per (SNR, modulation) pair, leading to a total of $175\,000$ examples, each of sequence length 1024. The task is to classify each sample in one of the seven modulation classes. We use 10\% of the dataset for validation and 10\% for testing. Cross-Entropy is used as the learning loss, while the test performance metric is classification accuracy.\\

The hyperparameters used for each of these tasks are grouped in table \ref{tab:tasks_params}. Note that for AugMod, we trained two models of different capacity, labelled \textit{AugMod - Large} and \textit{AugMod - Small}. We do not apply regularization to the small version of the model, as this results in a larger accuracy loss.

\begin{table}
\center
\caption{Hyperparameters for the different tasks. See equation \ref{eq:lambda_definition} and algorithm \ref{alg:hlra} for details.}
\begin{tabular}{lcccccc}
\hline
     &  $\bar{\lambda}$ & $R$ & $T_1$ & $T_2$ & $N$ & Epo.\\
\hline
SeqMNIST & $10^{-4}$ & 40 & 10 & 120 & 20 & 150 \\
DOCC10   & $10^{-3}$ & 20 & 5 & 25 & 10 & 50 \\
AugMod - L & $10^{-4}$ & 20 & 10 & 30 & 15 & 50\\
AugMod - S & 0 & 10 & n.a. & n.a. & 15 & 100\\\hline
\end{tabular}

\label{tab:tasks_params}
\end{table}

\subsection{Model's architecture}
\label{subsec:architecture}

Our model consists of a $l$-layered GRU with hidden-size $h$ and dropout applied after each layer, followed by a ReLU~(Rectified Linear Unit~\cite{nair2010rectified}) activation. At the end, a max pooling layer collapses the time dimension, producing a one-dimensional vector which is finally fed to a fully connected layer. Applying a pooling layer over the time dimension has the advantage of making the model invariant to the number of samples in the input signal\cite{AugMod}. Table~\ref{tab:archs_params} summarizes the hyperparameters characterizing the architectures of the models.

Note that for the SequentialMNIST task the model is slightly different. We remove the pooling layer, as a result of which we obtain a sequence as output.

\begin{table}
\center
\caption{Details of the model's hyperparameters characterizing the architecture.}
\begin{tabular}{lcccc}
\hline
     &  Input & Output & $l$ & $h$\\
\hline
SeqMNIST & 1 & 1 & 2 & 100 \\
DOCC10   & 1 & 10 & 3 & 150 \\
AugMod - Large & 2 & 7 & 3 & 150\\
AugMod - Small & 2 & 7 & 2 & 62\\\hline
\end{tabular}

\label{tab:archs_params}
\end{table}
\section{Results and Discussion}
\label{sec:results}

For each model architecture listed in Table~\ref{tab:archs_params}, we train two models: one baseline trained without any training adaptation (standard training), and another trained using nuclear regularization and HLRA (LRA-aware training). We denote these models \textit{Base} and \textit{LRA} respectively.

In this section we first present a qualitative analysis of the impact of training adaptation on the models' singular values. Then we show how much we can compress models using our method. Finally we discuss the size-accuracy trade-off permitted by Rank-Tuning.

\subsection{Impact of training adaptations on singular values}

\begin{figure}
    \begin{center}
    \includegraphics[width=\columnwidth]{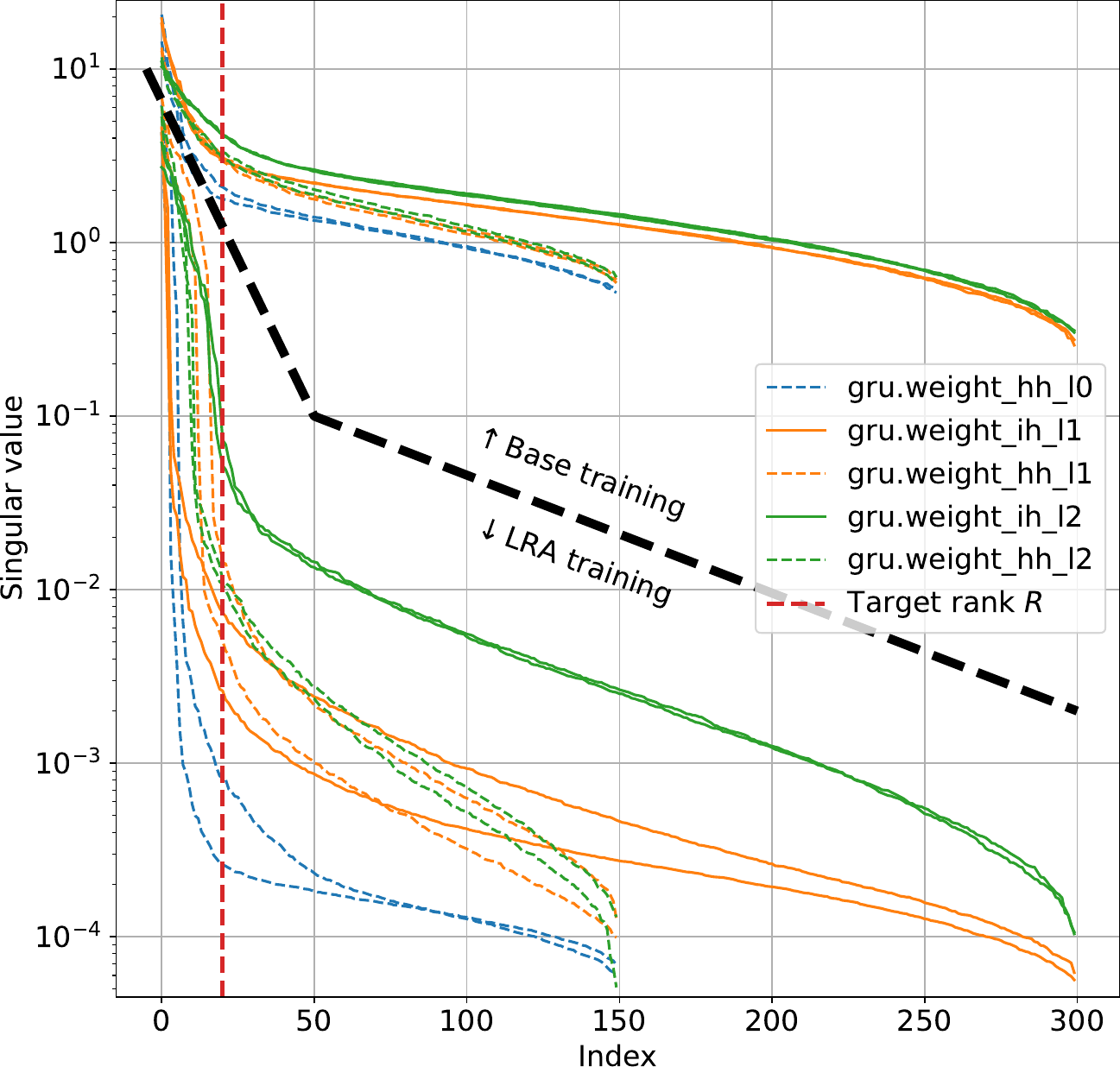}
    \caption{Singular values evolution, with and without training adaptations, for the different matrices. Curves above and under the black dashed line correspond to a \textit{Base} model and a \textit{LRA} model, respectively. GRUs being bilateral, each curve is present twice, one for each direction.}
    \label{fig:singular_values}    
    \end{center}
\end{figure}

Both nuclear regularization and HLRA are supposed to act on the singular values of the matrices. We can check that it is indeed the case by plotting the singular values in decreasing order, that is, the list of values $\sigma_1,...,\sigma_m$, which are the parameters of the diagonal of matrix $\Sigma$ (see equation~\ref{eq:svd_definition}). Figure~\ref{fig:singular_values} is an example of such a figure for the DOCC10 task, using a logarithmic scale.

We observe that for the \textit{Base} model, singular values decrease slowly from $\approx20$ to $\approx0.1$. This is in sharp contrast with the singular values of the \textit{LRA} model, which decrease steeply from $\approx7$ to $\approx10^{-4}$.  This clearly show the impact of training adaptations.

We can see that some matrices have singular values that decrease less rapidly than others. Indeed, the deeper the layer, the slower singular values decrease. We believe that matrices farther from input need more independent parameters to process the data they receive, their rank is therefore higher and this is reflected by their singular values. As a matter of fact, when we run the Rank-Tuning algorithm on the models, we observe that for each matrix, the selected rank is correlated with the decrease of the associated curve. The lower the rank, the faster the decrease. Thus, looking at this figure, one can quickly identify which matrices will be hard to compress (meaning that they have a high rank).

\subsection{Compression of neural networks}

As we mentioned in section~\ref{sec:experiments}, each dataset is split into three sets (train, validation and test). We use these to define the following compression procedure:
\begin{enumerate}
    \item train the model using the training set. We use the validation set to do early-stopping;
    \item after training, use the validation set to search for optimal factorization ranks;
    \item use the test set to measure performance of the uncompressed model;
    \item use the test set to measure performance of the compressed model.
\end{enumerate}

It turns out that the Rank-Tuning algorithm significantly outperforms\footnote{Meaning that models can be compressed further for an equivalent performance degradation.} the simpler rank-selection strategies that we exposed in section~\ref{subsubsec:simple_strategies}, at the cost of a higher computational cost. In the rest of this section, all the models are compressed using Rank-Tuning unless stated otherwise.

\begin{table*}

\center
\caption{Results on the different tasks. The FLOP metric is representative of the model's computational complexity and is normalized with respect of the number of samples processed ; ReLU and pooling are not taken into account. \textit{Before} and \textit{After} columns report metrics before and after compression respectively. Performance reduction is computed relatively to those of the \textit{Base} uncompressed model.}

\begin{tabular}{m{2.9cm}c|cc|ccc|ccc}
\hline
\multicolumn{1}{c}{\multirow{2}{*}{Task}}  & \multirow{2}{*}{Training} & \multicolumn{2}{c|}{FLOP/sample}     & \multicolumn{2}{c}{Performance} & \multirow{2}{*}{\begin{tabular}[c]{@{}c@{}}Perf.\\ reduction\end{tabular}} & \multicolumn{2}{c}{Parameters} & \multirow{2}{*}{\begin{tabular}[c]{@{}c@{}}Comp.\\ rate\end{tabular}} \\
\multicolumn{1}{c}{}                       &                           & Before                & After & Before          & After         &                                                                              & Before                 & After &                                                                       \\ \hline
\multirow{2}{*}{SeqMNIST $\downarrow$}     & Base                      & \multirow{2}{*}{248k} & 222k  & 0.303           & 0.306         & 1\%                                                                          & \multirow{2}{*}{243k}  & 218k  & 10\%                                                                  \\
                                           & LRA                       &                       & 80k   & 0.303           & 0.303         & 0\%                                                                          &                        & 76k   & \textbf{69\%}                                                         \\ \hline
\multirow{2}{*}{DOCC10 $\uparrow$}         & Base                      & \multirow{2}{*}{964k} & 224k  & 79.1            & 78.2          & 1.1\%                                                                        & \multirow{2}{*}{954k}  & 214k  & 78\%                                                                  \\
                                           & LRA                       &                       & 75k   & 78.7            & 78.3          & 1\%                                                                          &                        & 65k   & \textbf{93\%}                                                         \\
\multirow{2}{*}{AugMod - Large $\uparrow$} & Base                      & \multirow{2}{*}{964k} & 368k  & 85.0            & 83.8          & 1.4\%                                                                        & \multirow{2}{*}{954k}  & 359k  & 62\%                                                                  \\
                                           & LRA                       &                       & 116k  & 84.9            & 84.3          & 0.8\%                                                                        &                        & 106k  & \textbf{89\%}                                                         \\
\multirow{2}{*}{AugMod - Small $\uparrow$} & Base                      & \multirow{2}{*}{98k}  & 75k   & 83.8            & 83.0          & 1\%                                                                          & \multirow{2}{*}{95k}   & 72k   & 25\%                                                                  \\
                                           & LRA                       &                       & 23k   & 82.8            & 82.7          & 1.3\%                                                                        &                        & 20k   & \textbf{79\%}                                                         \\ \hline
\end{tabular}
\label{tab:results}
\end{table*}

Results on the different tasks are regrouped in Table~\ref{tab:results}. For a given task, we report the size (number of parameters), FLOP~(Floating-point Operations), and performance of \textit{Base} and \textit{LRA} models, before and after applying the compression step (i.e. the Rank-Tuning algorithm). As we can see, our training adaptation method significantly improves the compressibility of the \textit{LRA} models. With a performance degradation of roughly 1\%, we achieve very high compression rates, from 69\% to 93\% depending on the task, whereas for the \textit{Base} models (standard training), the compression rates range from 10\% to 78\%. Note that we systematically obtain better compression rates when using training adaptations.

We observe that compressing the models almost always leads to a small performance reduction. This is expected: since compression is not lossless, we lose some information (parameters) when factorizing matrices. However, by design these are unimportant parameters, that's why we are able to achieve high compression rates without hurting much the performance.
\noindent We also observe on Table~\ref{tab:results} that the use of training adaptations can lead to a performance reduction as high as one accuracy point \textit{before} compression. If this is statistically significant, this would mean that LRA-aware training slightly hurts the model's performance in order to improve its compressibility. In order to disentangle this observation from the natural variance due to weights initialization, we trained 10 \textit{Base} and 10 \textit{LRA} models on the SequentialMNIST task and checked their MSE. We obtained a mean MSE of $0.321 (\sigma = 0.022)$ for \textit{Base} and of $0.317 (\sigma = 0.014)$ for \textit{LRA} models respectively, where $\sigma$ gives the standard deviation. We thus observe no significant differences between the two training methods. Though we did not run the same experiment with the other tasks due to the high computational cost, we believe we would observe a similar trend.

Note that we don't report inference speed because we're unable to measure it in a satisfactory manner. Although in theory network compression via matrix factorization decreases both the model's size and inference time (as explained in section~\ref{subsec:svd_basics}), in practice inference relies heavily on matrix multiplications optimized by linear algebra libraries. Thus, without an implementation of a two-step multiplication as described in equation~\ref{eq:matrix_multi}, it is not possible to take advantage of the factorization at inference time. Nonetheless, we report the FLOP metric in Table~\ref{tab:results} to give an indication of the speed gain that could be achieved using an adequate implementation.

\subsection{Size-accuracy trade-off}
\label{subsec:trade-off}

\begin{figure}
    \begin{center}
    \includegraphics[width=\columnwidth]{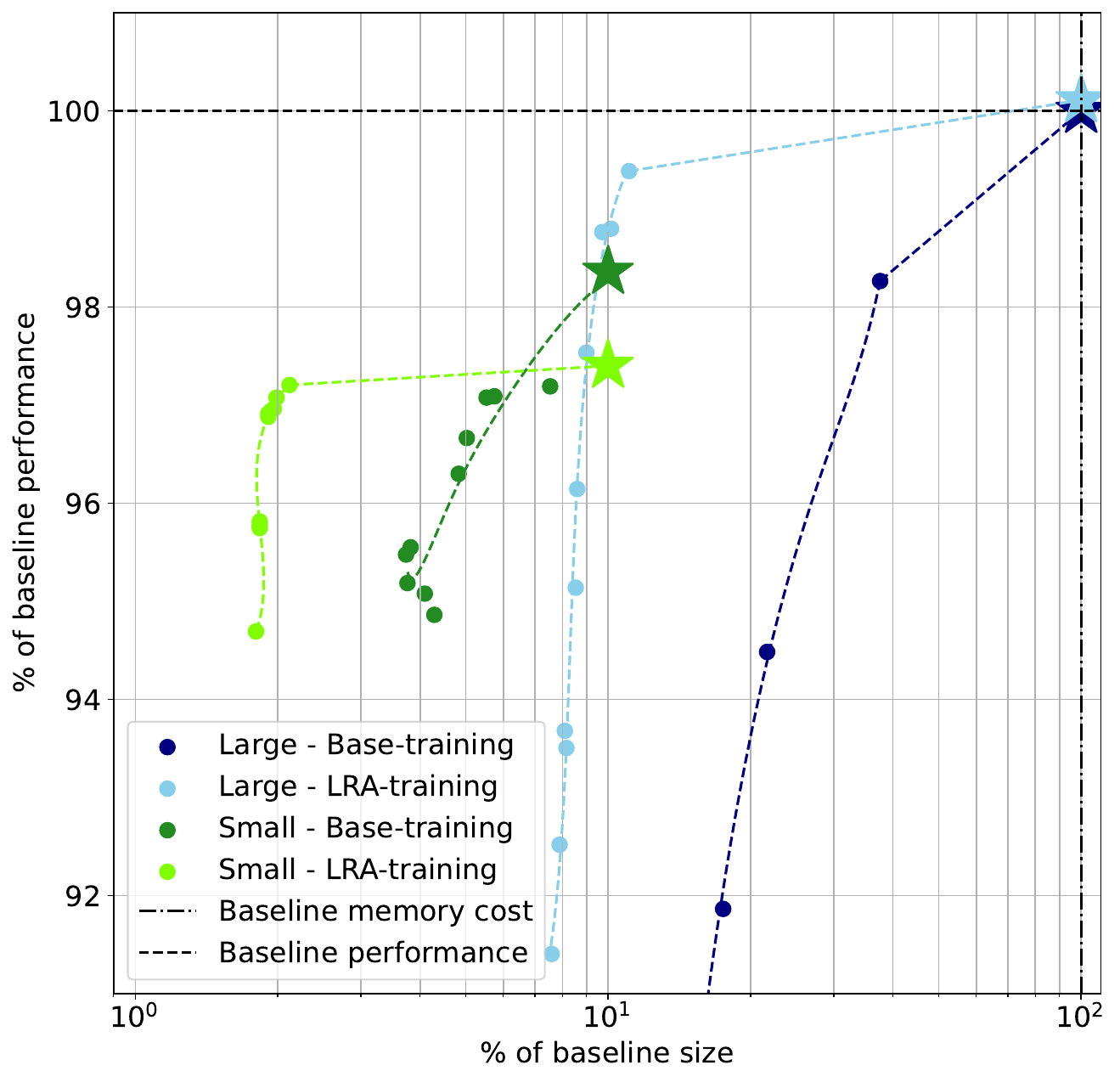}
    \caption{A representation of the size-accuracy trade-off for the AugMod task. Uncompressed and compressed models are represented by stars and dots, respectively.}
    \label{fig:size_accuracy_trade_off}    
    \end{center}
\end{figure}

The Rank-Tuning algorithm~\ref{alg:rank_tuning} is parameterized by the precision tolerance $\Delta_p$. By varying $\Delta_p$, we can produce several compressed models with different compression rates and performance degradation. Note that this can be done without additional computations, by slightly adapting the Rank-Tuning algorithm to use a range $[\Delta_1, \Delta_2, ...]$.

We trained several models (Large and Small architectures, with and without LRA-aware training) on the AugMod dataset, then compressed them with $\Delta_p = \frac{i}{1000}p^{*}$ for $1 \leq i \leq 10$. For each of these factorized models, we plotted its size and performance relative to the baseline (Large, uncompressed). The results are shown on figure~\ref{fig:size_accuracy_trade_off}. As we can see, LRA-aware training significantly improves over standard training both in terms of compression rate and performance. Moreover, this graph clearly shows one of the key advantage of our method: one can choose the desired size-accuracy trade-off according to what is needed for the model's deployment.

Finally, we discuss one key question in model optimisation: in terms of performance, is it better to use a large compressed model or a small uncompressed one? According to our experiments on the AugMod task, a large model trained with LRA-aware training can be compressed to the size of a small model, while retaining superior accuracy (see figure~\ref{fig:size_accuracy_trade_off}, skyblue curve). However, if we want to obtain a model as small as possible, it is better to compress a small model. On the AugMod dataset, we were able to compress the small model nearly $50\times$ compared to the large model, while retaining 97\% of its accuracy.
\section{Conclusion}
\label{sec:conclusion}

Compressing a neural network while maintaining its performance remains challenging. In this work, we used regularization and training-time factorization to train recurrent neural networks which can be more effectively compressed. Using these training adaptations in conjunction with a post-training rank-selection, we were able to reach very high rates of compression with no or little performance loss. Our experiments show that the proposed method can be used to select a good trade-off between compression and performance. We leave as future work the application of this method to other architectures, as well as the exploration of the combination of low-rank approximations with other compression methods like quantization or distillation.

\bibliographystyle{IEEEtran}
\bibliography{mybib}

\end{document}